# A Unified Taxonomy of Deep Syntactic Relations

**Kira Droganova**
Charles University
Faculty of Mathematics and Physics
Praha, Czechia
droganova@ufal.mff.cuni.cz

**Daniel Zeman**
Charles University
Faculty of Mathematics and Physics
Praha, Czechia
zeman@ufal.mff.cuni.cz

## Abstract

This paper analyzes multiple deep-syntactic frameworks with the goal of creating a proposal for a set of universal semantic role labels. The proposal examines various theoretic linguistic perspectives and focuses on Meaning-Text Theory and Functional Generative Description frameworks.

For the purpose of this research, data from four languages is used – Spanish and Catalan (Taulé et al., 2011), Czech (Hajič et al., 2017), and English (Hajič et al., 2012).

This proposal is oriented towards Universal Dependencies (de Marneffe et al., 2021) with a further intention of applying the universal semantic role labels to the UD data.

## 1 Introduction

Linguistic research and multilingual natural language processing need annotated data in many languages, ideally following a uniform annotation framework. For morphology and surface syntax, Universal Dependencies (UD)[1] (de Marneffe et al., 2021) is the current de-facto standard of such a framework. Nevertheless, despite being an important linguistic resource, UD is only one step towards natural language understanding. The mapping between surface syntax and meaning is not straightforward, as the same meaning can be encoded in various syntactic constructions (e.g., active vs. passive clauses), and vice versa, one syntactic construction can be used to convey different meanings (e.g., the English preposition *on* can express location, time, or other verb-specific roles as in *I rely on him*). Therefore there are datasets that attempt to annotate another layer (or multiple layers) of the language, which is closer to the meaning and is variously termed 'deep-syntactic', 'tectogrammatical', or even 'semantic'. Unfortunately, the annotations in this layer have not reached the level of cross-linguistic uniformity and interoperability that UD set for morphology and surface syntax.

Deep-syntactic annotation can cover a variety of phenomena but in the present paper, we focus on the inventory of deep-syntactic (or semantic) relations between words. We study the inventories used in existing annotation frameworks, compare them and propose a unified inventory where the same meaning would have the same label across datasets. This unified set of relations should be applicable to any language. Ideally, it should be possible to map relations from existing frameworks onto this inventory without loss of information; while there is no guarantee that this ideal goal is achievable, we want to get as close to it as possible.

There are two related projects worth mentioning here. Universal Proposition Bank (Jindal et al., 2022) provides semantic role annotation for 23 languages, based on their UD treebanks. As the name suggests, semantic role labels follow the PropBank (Kingsbury and Palmer, 2002). Second, a recent proposal by Evang (2023) defines the CRANS annotation scheme to annotate semantic roles on top of UD. Only a few coarse and cros-linguistically applicable valency frames (superframes) are defined in CRANS in order to avoid reliance on large-coverage language-specific valency dictionaries.

The paper is organized as follows. We first (Section 2) give a brief overview of annotation frameworks that we considered for this study, explaining how we selected the ones to focus on in the rest of the paper. In Section 3, we survey the deep-syntactic relations in Meaning-Text Theory, in Section 4 we do the same with Functional Generative Description. Finally, in Section 5 we propose a unified set of relations to which the other two can be mapped.

[1] https://universaldependencies.org/

## 2 Selection Criteria

To be able to work with the most relevant frameworks, the framework selection criteria and terminology proposed in (Žabokrtský et al., 2020) is re-used and adapted. Although the purpose of this survey is to provide an overview of how sentence meaning is represented in selected deep-syntactic frameworks, it also provides both a direct comparison to UD and also suggestions for a discussion on a unifying approach to sentence meaning. This makes it a natural point of departure for this research by creating a proposal for universal semantic role labels.

This work is directly related and built upon UD. It is therefore focused on meaning representations whose backbone structure can be described as a graph over words (possibly with added non-lexical nodes) corresponding to entities, events, properties, or circumstances, with edges representing meaningful relationships among them. Basic UD is the surface sentence representation. Thus, approaches that handle only the original sequence of word forms and do not make any abstraction above overt morphological, lexical, or syntactic means are not included.

### 2.1 Basic Criteria

Similar to Žabokrtský et al. (2020), this research only includes frameworks capable of analyzing whole authentic sentences of natural languages. Purely lexicographical approaches are not included.

The paper examines approaches that have been extensively studied for a longer period of time, and have been utilized in natural language applications.

Another important criterion addresses data availability – whether a framework has enough data, meaning that the framework has a publicly available associated corpus of a reasonable size or is available in the main part of the Linguistic Data Consortium catalog.

The pre-selection of frameworks including associated corpora and major lexicographical resources was taken from Žabokrtský et al. (2020) (Table 1) and further refined into a short list of deep-syntactic frameworks that was used for the proposal.

### 2.2 Additional Criteria

As it was mentioned in Section 2, this proposal for universal semantic role labels is oriented towards UD, meaning that the proposal must be built upon existing guidelines and take into account UD specific features:

- **Representation:** Basic representation and Enhanced representations (Schuster and Manning, 2016) are available in UD (Zeman, 2021), although for the majority of languages the enhanced representation still has to be generated automatically (Droganova and Zeman, 2019). The two graphs are stored in the same file side-by-side. In many cases the basic tree is a subset of the enhanced graph, but it is not guaranteed.[2] The universal deep-syntactic relations must extend the enhanced representation without breaking it.

- **Content vs. Function Words:** Regardless of the chosen representation, in UD, function words have nodes of their own, but they are attached to content words and treated like their attributes.

- **Data structure:** The enhanced UD graph is typically quite close to the rooted tree of the basic representation – the two structures can be identical. Propagation of dependencies across coordination and propagation of arguments of control verbs may cause the tree to become a directed acyclic graph (DAG). Cycles appear only in sentences with relative clauses. The enhanced UD graph can also include empty (copied) nodes for deleted predicates so their arguments and adjuncts are reasonably attached to the graph.

Taking a closer look at the frameworks listed in Table 1 it turns out that some of the frameworks do not satisfy all the criteria listed above or match UD-specific features, thus they were not included in this research:

- **Paninian Framework:** This framework defines 6 karaka relations (Bharati et al., 1996). The relations are very coarse-grained and do not directly correspond to semantic roles. However, the main reason for not including this framework is data availability.[3]

- **Enju Predicate–Argument Structures:** Enju structures distinguish three types

[2] Sometimes a basic edge is omitted from the enhanced graph.

[3] At the time of writing this paper, the data was not available in LDC or any open-access repository.

| Framework | Associated corpus | Lexical resource | Languages |
|---|---|---|---|
| 1. Paninian framework | HDTB, UDTB | | hi, ur, bn, te |
| 2. Meaning-text theory (MTT) | SynTagRus, AnCora-UPF | ECD | ru, en, es, fr |
| 3. Functional Generative Description (FGD) | PDT, PCEDT | PDT-VALLEX | cs, en |
| 4. PropBank | PropBank + NomBank + PDTB | PropBank lex. | en, ar, zh, fi, hi, ur, fa, pt, tr, de, fr |
| 5. FrameNet-based approaches | | FrameNet | en, de, fr, ko |
| 6. Enju | Enju Treebank | | en, zh |
| 7. DELPH-IN | DeepBank | ERG | en, de, es, ja |
| 8. Sequoia | Sequoia | | fr |
| 9. Abstract Meaning Representation (AMR) | AMR Bank | PropBank lex. | en, zh, pt, ko, vi, es, fr, de |
| 10. Universal Conceptual Cognitive Annotation (UCCA) | English Wiki, parallel fiction, etc. | | en, de, fr |
| 11. Enhanced Universal Dependencies | Universal Dependencies | | ar, bg, cs, en, et, fi, it, lt, lv, nl, pl, ru, sk, sv, ta, uk |

Table 1: Selection of frameworks including associated corpora and major lexicographical resources from (Žabokrtský et al., 2020)

of arguments: ARG1 (semantic subject), ARG2 (semantic object), and MODARG (modifier) (Yakushiji et al., 2005). These types of arguments are too coarse-grained for the purpose of this research.

- **Abstract Meaning Representation:** In AMR, sentences are represented as directed graphs that treat non-leaf nodes as variables and only leaf nodes are labeled with concepts (Banarescu et al., 2013). Nodes in an AMR graph are unordered; Any correspondence between nodes and surface strings is hidden by design, making mapping to surface representation extremely unreliable.

- **FrameNet-Based Approaches:** In this approach semantic relations in a sentence are represented using the FrameNet semantics framework (Fillmore, 1976; Fillmore and Baker, 2001). The FrameNet project is a lexical database of English based on examples of how words are used in actual texts – it consist of frame elements whose labels are chosen with regard to the particular situation making the labels extremely fine-grained and not practical for this proposal. Although FrameNet-like databases have been built for a number of languages,[4] aligning the FrameNets across languages is work in progress.

- **Universal Conceptual Cognitive Annotation (UCCA):** UCCA graphs distinguish terminal and non-terminal nodes (Abend and Rappoport, 2013). Terminal nodes are anchored in the surface text making words and multi-word chunks their labels. Non-terminal nodes do not have labels and can be characterized in terms of the categories of its outgoing edges. Edges are labeled with 12 coarse-grained categories, such as P – Process, A – Participant, D – Adverbial, E – Elaborator, and N – Connector. The way the categories were designed makes them quite difficult to map to other frameworks. For instance, the Participant label includes both Agent/arg0 and Patient/arg1, making these categories indistinguishable.

In view of considerations above the following short list of deep-syntactic frameworks was created:

1. Meaning-Text Theory (Žolkovskij and Mel'čuk, 1965)
2. Functional Generative Description (Sgall, 1967)
3. PropBank (Kingsbury and Palmer, 2002)
4. Sequoia (Candito et al., 2014)

In the present paper we focus on the perspective of the Meaning-Text Theory and Functional Generative Description, leaving the comparison to the other two frameworks for future work.

## 3 Meaning-Text Theory

### 3.1 Overview

The goal of the Meaning-Text Theory (MTT) is to write systems of explicit rules that express the correspondence between meaning and text (or sound) in various languages (Kahane, 2003). The Meaning-Text approach to language was put forward in the framework of research in machine translation in the early 1960s (Žolkovskij and Mel'čuk,

[4]The FrameNet web page (https://framenet.icsi.berkeley.edu) mentions French, Chinese, Brasilian Portugese, German, Spanish, Japanese, Swedish, and Korean

1965) and since then has been extensively worked on. The correspondence between meanings and texts is completely modular. MTT defines a seven-level representation that describes the relation between form and meaning:

- surface-phonological representation (text)
- deep-phonological
- surface-morphological
- deep-morphological
- surface-syntactic
- deep-syntactic
- semantic representation (meaning)

MTT utilizes dependency, which means that on the deep-syntactic level the structure of a sentence corresponds to a rooted directed acyclic graph[5] where nodes correspond to content words and can be ordered following the surface word order, and edges represent dependency relations between nodes. A deep-syntactic graph may contain copied nodes that are used to represent controlled subjects or elliptic constructions. The set of relations used in MTT is rather coarse-grained. It consist of a set of numbered arguments expressing their degree of proximity to the predicate, and "utility" relations such as `ATTR` for attributes and other modifiers, `COORD` for coordination, and `APPEND` for parentheses, interjections, and other similar items. MTT can be characterized by the massive relocation of syntactic information into the lexicon. The lexicon in MTT is represented by an explanatory combinatorial dictionary (ECD) (Mel'čuk, 2006), which includes entries for all of the lexical items of a language along with information on their combinatorics and specific rules. Lexical relations among lexemes in the lexicon are captured by Lexical Functions (LF).

The MTT scheme is applied in a corpus for Russian (SynTagRus) and two treebanks for Spanish and Catalan (AnCora 2.0). Although SynTagRus (Apresjan et al., 2006) contains morphological annotation, surface-syntactic dependency trees, lexical semantic and lexical-functional annotation, the deep-syntactic and semantic annotation seems unavailable yet. Therefore, this research only considers data from AnCora.

[5]But the structure can contain cycles in case of coreference.

### 3.2 Available Data

UD_Spanish-AnCora and UD_Catalan-AnCora are two available treebanks that provide original semantic role labels. The original annotation was done in a constituency framework as a part of the AnCora project (Taulé et al., 2008). The corpora were converted to dependencies and used in the CoNLL 2009 shared task (Hajič et al., 2009) and later converted to Universal Dependencies (Martínez Alonso and Zeman, 2016). Table 2 shows the number of sentences and tokens for each corpus.

The two corpora consist mainly of newspaper texts annotated at morphological, syntactic, and semantic levels. Table 5 shows 20 thematic role labels and their frequencies. The frequencies are similar in both treebanks (see Figure 1 and Figure 2). Each label can be combined with argument position (see Description in Table 5 in Appendix B). The arguments required by the verb sense are incrementally numbered, expressing their degree of proximity in relation to its predicate (Palmer et al., 2005). There are seven possible argument slots: `arg0`, `arg1`, `arg2`, `arg3`, `arg4`, `argM` and `argL`, where adjuncts are tagged with `argM` and lexicalized complements of light verbs are marked with `argL`.

### 3.3 Thematic Roles

Taulé et al. (2011) describe a set of 20 thematic roles in detail. Each of the roles can be mapped to several syntactic functions and argument positions.

The roles are based on Lexical Semantic Structure (LSS) – the concept defined assuming lexical decomposition (Levin and Rappaport Hovav, 1994; Rappaport Hovav and Levin, 1998). LSS determines the number of arguments that a verbal predicate requires and the thematic roles of these arguments, and describes the syntactic functions of the arguments. Each LSS restricts the set of all possible diatheses and each verb sense is associated to one LSS. Diatheses must be understood as the syntactic expression of a semantic opposition (Taulé et al., 2011).

The subsections below provide a closer look at the AnCora data and show examples of each role in different syntactic positions.[6] The examples are taken from the Spanish data; the data for Spanish and Catalan is of the same origin and has a similar

[6]The examples are selected according to the frequency of their syntactic labels.

| Set | UD_Spanish-AnCora, sent./tok. | UD_Catalan-AnCora, sent./tok. |
|---|---|---|
| Train | 14,287/459,237 | 13,123/434,140 |
| Development | 1,654/54,220 | 1,709/58,795 |
| Testing | 1,721/54,437 | 1,846/60,107 |

Table 2: **UD_Spanish-AnCora, sent./tok.:** the number of sentences/tokens in the UD Spanish-AnCora treebank; **UD_Catalan-AnCora, sent./tok.:** the number of sentences/tokens in the UD Catalan-AnCora treebank.

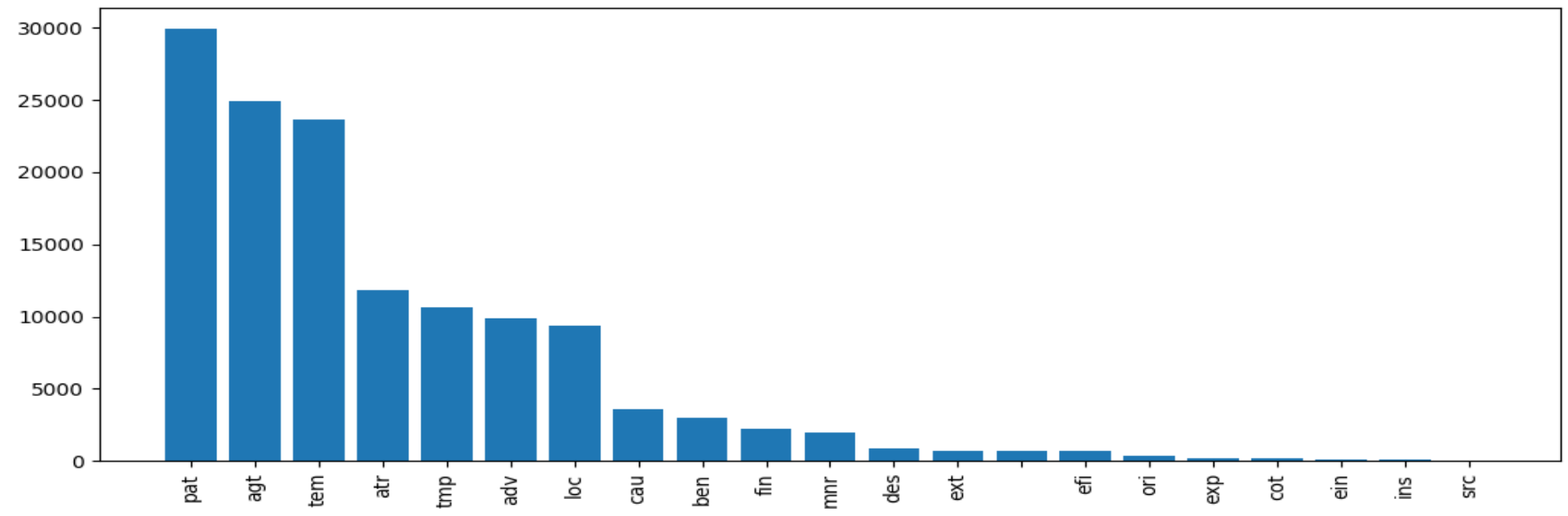

Figure 1: Label frequency in Catalan data.

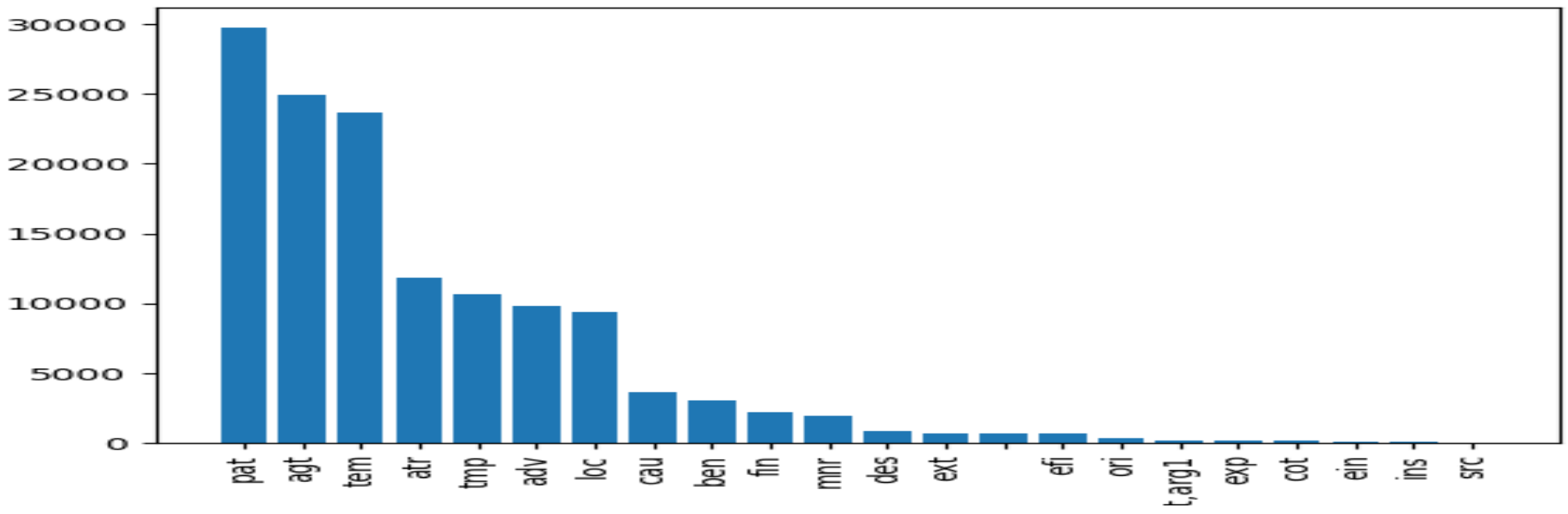

Figure 2: Label frequency in Spanish data.

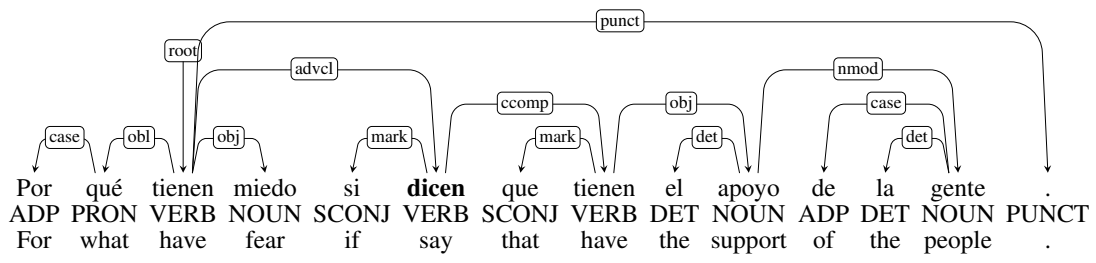

Figure 3: An example of `ArgM:adv` – advcl.

role label distribution (see Table 5).

### Adverbial: adv

The Adverbial role is a broad category that corresponds to non-specific adjuncts and can be expressed by the UD syntactic relations *advcl* (Figure 3), *advmod* or *obl*.

### Agent: agt

The Agent role is associated with the external causer argument that is expressed as the syntactic subject. In some cases the external argument (`arg0`) may be expressed as an oblique agent complement, keeping its original Agent role as well. The Agent role can be expressed syntactically as *nsubj*, *det* (Figure 4), *nmod*, and *obl* (Figure 5).

### Attribute: atr

The Attribute role refers to the third position (arg2) position in the state-attributive LSS that is typically expressed as the direct object. Other examples that can be found in the data are *root* (Figure 6) and *advcl* (Figure 7)

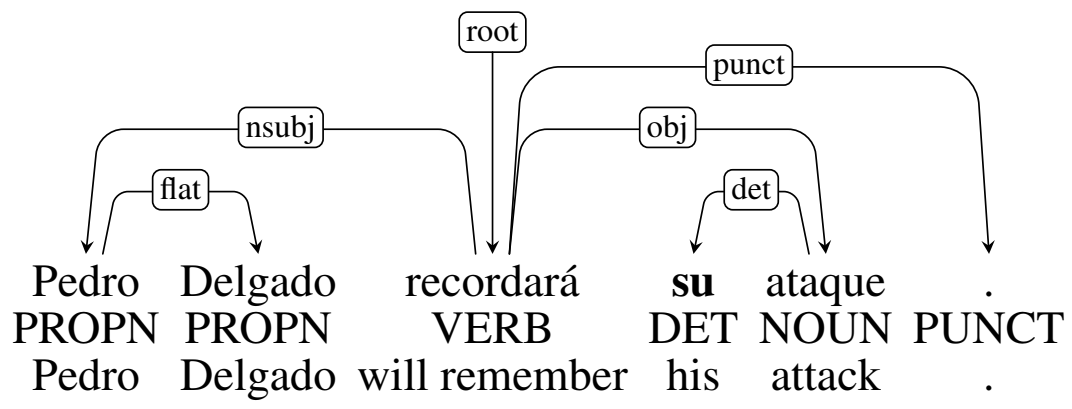

Figure 4: An example of `Arg0:agt` – det.

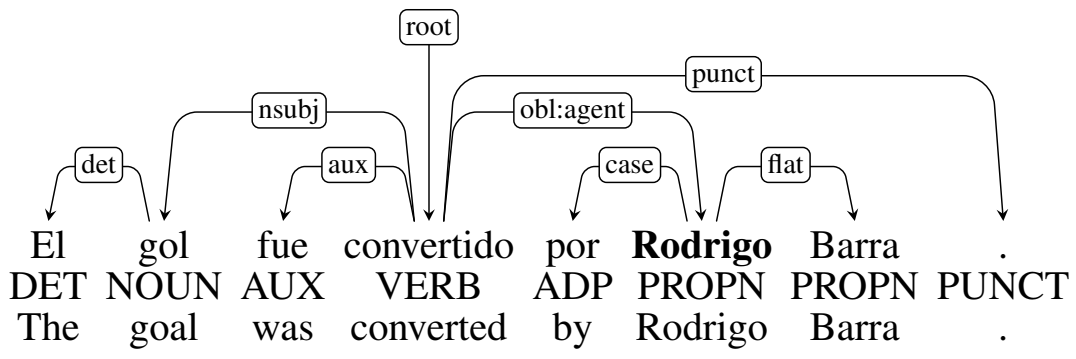

Figure 5: An example of `Arg0:agt` – obl.

## Beneficiary: ben

The Beneficiary role refers to the third argument (arg2) in the ditransitive-patient-benefactive LSS that is syntactically expressed as the indirect object.

## Cause: cau

The Cause role is a part of the transitive-causative LSS. Transitive-causative verbs associate the external causer argument (x) with the semantic predicate cause and the internal participant that undergoes the change with the argument (y). The argument x corresponds to the Cause role; It is syntactically the subject.

The Cause role can also take an adjunct position. In that case it receives *obl* or *advcl* (Figure 9) labels.

## Cotheme: cot

The Cotheme role refers to the third argument position (`arg2`) in the ditransitive-theme-cotheme LSS or the unaccusative-cotheme LSS. This role is expressed as a prepositional object – *nmod* or *obl* syntactic labels.

## Destination: des

The Destination role typically corresponds to the fifth argument position (`arg4`) that is most frequently expressed as *obl* and *nmod*.

## Experiencer: exp

In inergative-experiencer LSS, the Experiencer role refers to the first argument (`arg0`) that is expressed as the subject.

When the Experiencer role is a part of the state-experiencer LSS, it refers to the third argument (`arg2`). In this case it is expressed as the indirect object syntactically (Figure 10).

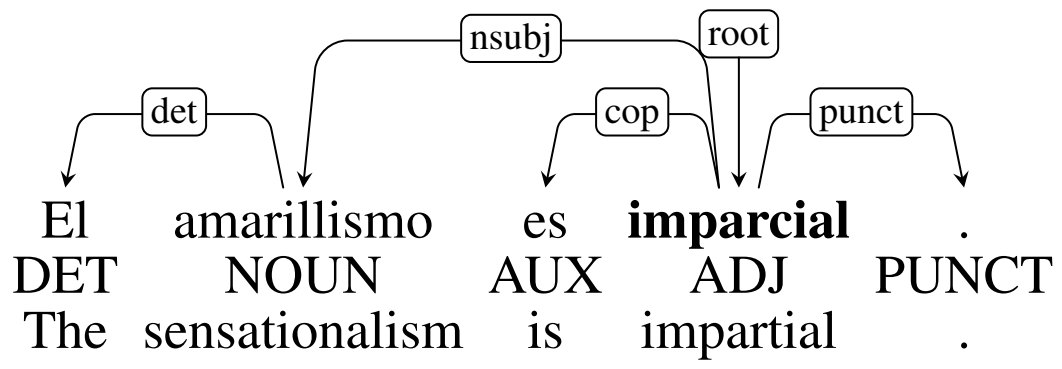

Figure 6: An example of `Arg2:atr` – root.

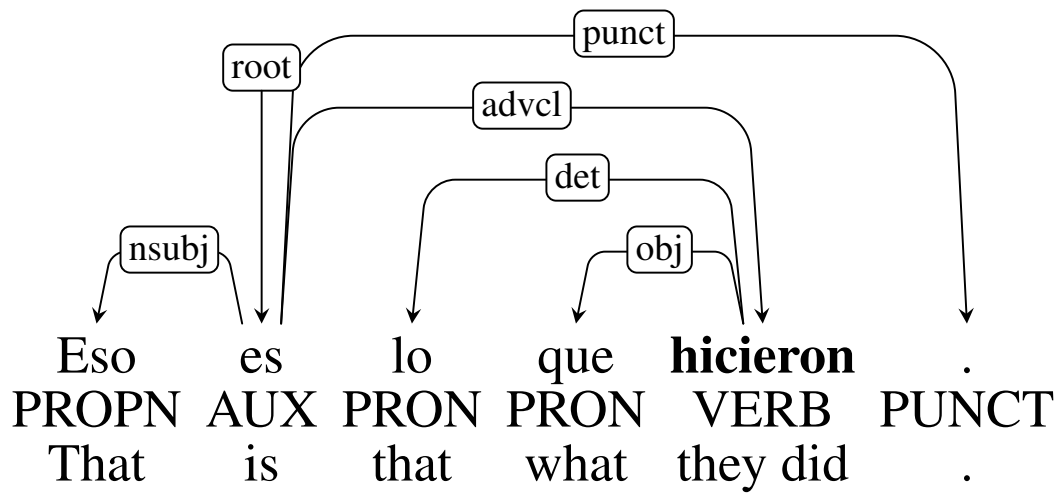

Figure 7: An example of `Arg2:atr` – advcl.

## Final State: efi

The Final State role refers to the third argument position (arg2) in the transitive-causative-state LSS or the unaccusative-state LSS. Arg2 can be expressed as an adjunct, a prepositional object or a predicative complement (Figure 11).

## Initial State: ein

The Initial State role is similar to the Final State role with the difference that it occurs in the data less frequently. It refers to the third argument position (`arg2`) in the transitive-causative-state LSS or the unaccusative-state LSS. Arg2 can be expressed as an adjunct, a prepositional object or a predicative complement.

## Instrument: ins

The Instrument role refers to the third argument position (`arg2`) in the transitive-causative-instrumental LSS (Figure 12)

## Location: loc

The Location role can occur in multiple LSS. However, in all of them, the Location role can be expressed as the third argument (`arg2`), typically a prepositional object or an adjunct (Figure 13) on the syntax level. Except for the ditransitive-theme-locative LSS, where this role can be expressed only as an adjunct.

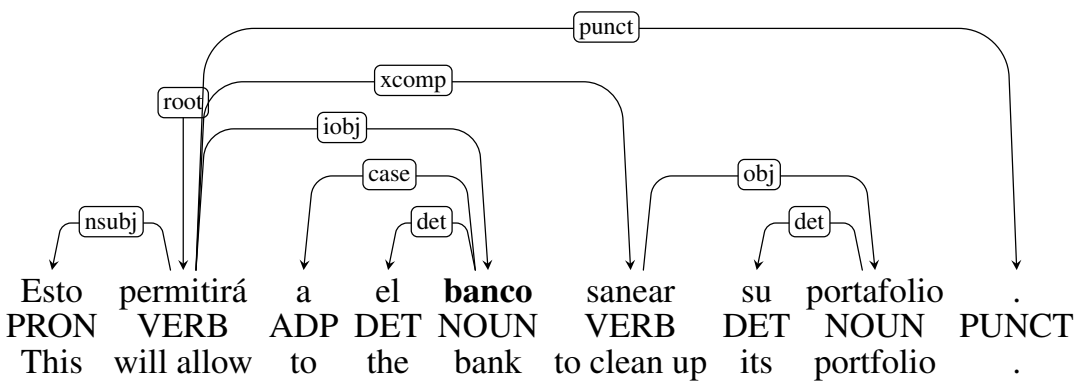

Figure 8: An example of `Arg2:ben` – iobj.

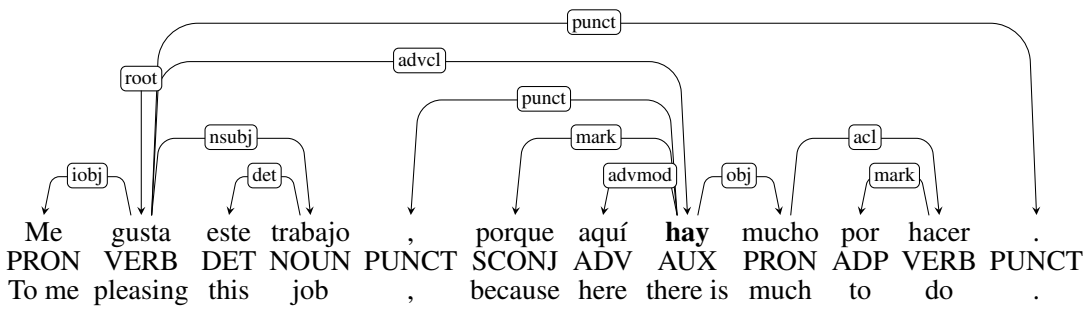

Figure 9: An example of `ArgM:cau` – advcl.

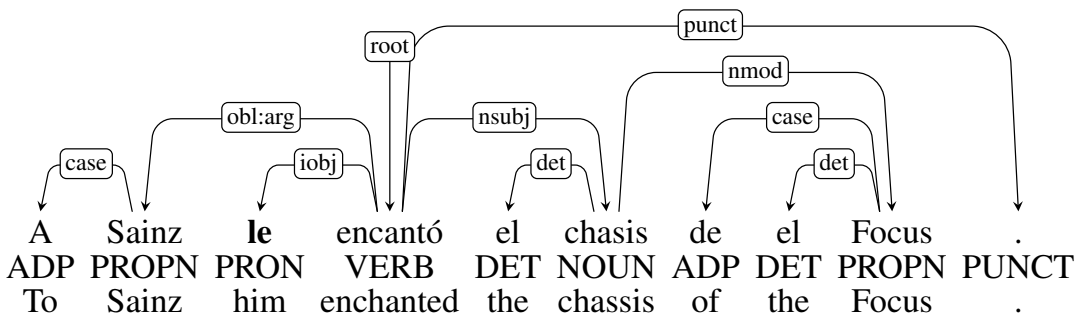

Figure 10: An example of `Arg2:exp` – iobj.

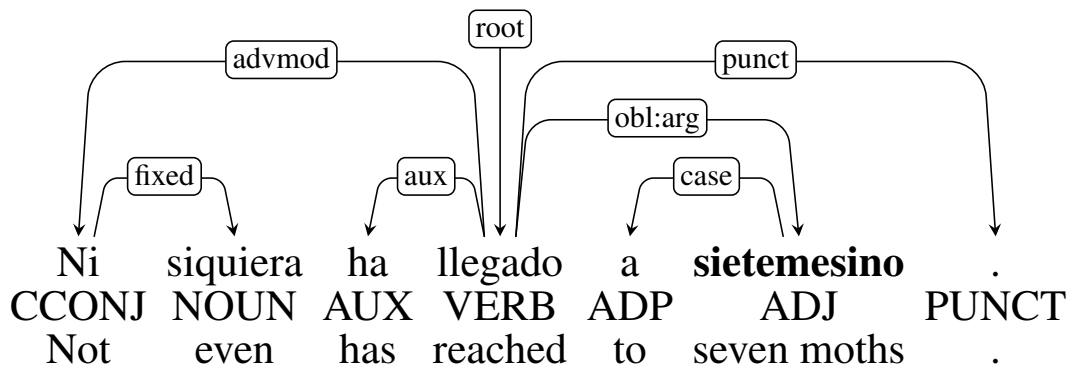

Figure 11: An example of `Arg2:efi` – obl.

In the ditransitive-patient-locative and the unaccusative-passive-ditransitive LSS, the semantic interpretation of Location is bound to a physical location in space. In the unaccusative-motion LSS the semantic interpretation of this role signifies a more specific destination or origin.

**Manner: mnr**

The Manner role refers to an adjunct (`ArgM`) that can receive one of the following syntactic labels: *obl* or *advmod* (Figure 14).

**Origin: ori**

The Origin role occurs in the data less frequently; It marks the place of origin and typically takes the fourth argument position (`arg3`). The most frequent syntactic label is *obl*.

**Patient: pat**

In the transitive-agentive-patient LSS and the ditransitive-agentive group of LSS, the Patient refers to the second argument position (`arg1`) that is expressed as the direct object. The Patient role refers to the second argument position (`arg1`) in the unaccusative-passive-ditransitive LSS and the unaccusative-passive-transitive. In both cases it is expressed as the syntactic subject.

**Purpose: fin**

The Purpose role refers to an adjunct; most frequently it is expressed as *advcl* on the syntactic level (Figure 15).

**Source: src**

The Source role refers to the first argument position (`arg0`) in the inergative-source LSS (Figure 16).

**Theme: tem**

The Theme role can occur in multiple LSS. In the transitive-causative group of LSS, the Theme role takes the second argument position (`arg1`). In the transitive-agentive LSS group, the Theme role also takes the second argument position (arg1), but its syntactic function is always as a prepositional object with the exception of the ditransitive-theme-cotheme LSS where its syntactic function is as the direct object. If the Theme role occurs in the state-attributive group of LSS, it refers to the second argument (arg1) that is syntactically the subject. When the Theme role occurs in the unaccusative-motion group of LSS as the second argument (`arg1`), it refers to the syntactic subject. If the Theme role refers to the third argument, it can be expressed as a prepositional object or an adjunct on the syntactic level. The Theme role most frequently receives one of the following syntactic labels: *nsubj*, *csubj*, *obj*, *nmod*, and *obl*.

**Time: tmp**

The Time role refers to temporal adjuncts that most frequently receive the following syntactic labels: *obl*, *advmod*, and *advcl*.

**Empty label: argL**

The argL label refers to lexicalized arguments of light verbs. This label does not receive any role label and most frequently occurs as *obl* or *obj* (Figure 17) on the syntactic level.

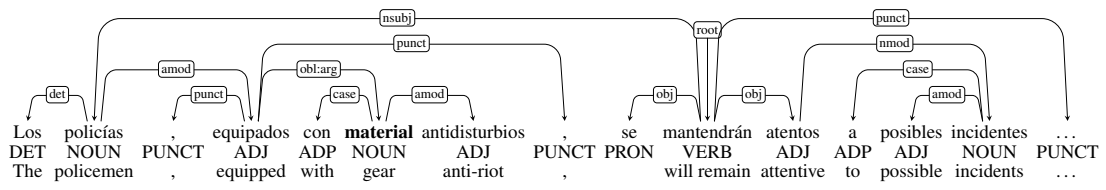

Figure 12: An example of `Arg2:ins` – obl.

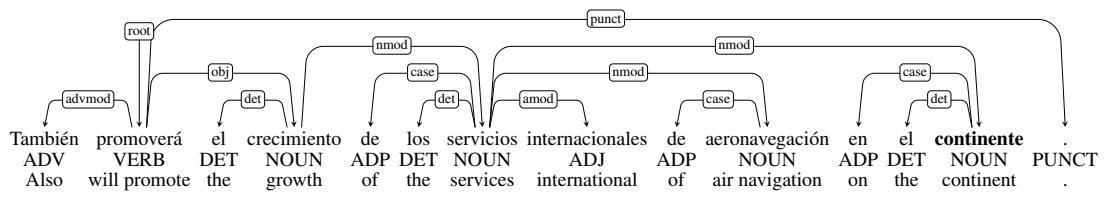

Figure 13: An example of `ArgM:loc` – nmod.

## 4 Functional Generative Description

### 4.1 Overview

Functional Generative Description (FGD) was introduced by Sgall (1967) in the beginning of 60's and has been gradually developed since then. FGD represents a dependency-based generative description that is based on a multilayer design reflecting the relation of form and function. Continuing the tradition of Prague School, special attention is paid to the phenomenon of topic–focus articulation.

FGD is a stratificational grammar formalism that treats the sentence as a system of interlinked layers:

- phonetic
- phonological
- morphemic
- analytical (surface syntax)
- tectogrammatical (deep syntax)

FGD is focused on the higher layers of the language description, from the morphemic one through the analytical to the tectogrammatical (deep-syntactic) layer that is considered the primary focus (Sgall et al., 1986).

The tectogrammatical representation describes the meaning of the sentence, thus synonymous sentences have a single representation on this level, while an ambiguous sentence has more than one tectogrammatical representation. The tectogrammatical layer contains complete information on the sentence required for its transduction on the lower layers.

Each sentence is represented as a dependency tree with labeled nodes and edges. Nodes represent the meaning units of the sentence containing their lexical and (deep) morphological information. Nodes in an FGD graph are ordered, which helps to capture the information structure of the sentence (topic-focus articulation). Edges stand for (deep) syntactic relations between the relevant nodes (Petkevič, 1995; Sgall et al., 1986). Deep-syntactic relations (functors) are linked to the valency lexicon which specifies which of the roles constitute the valency frame of the verb (being either obligatory or optional).

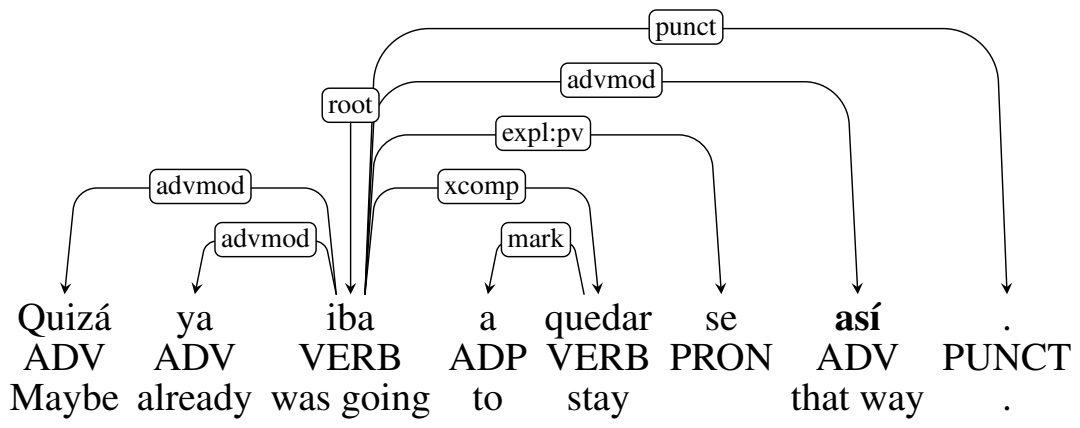

Figure 14: An example of `ArgM:mnr` – advmod.

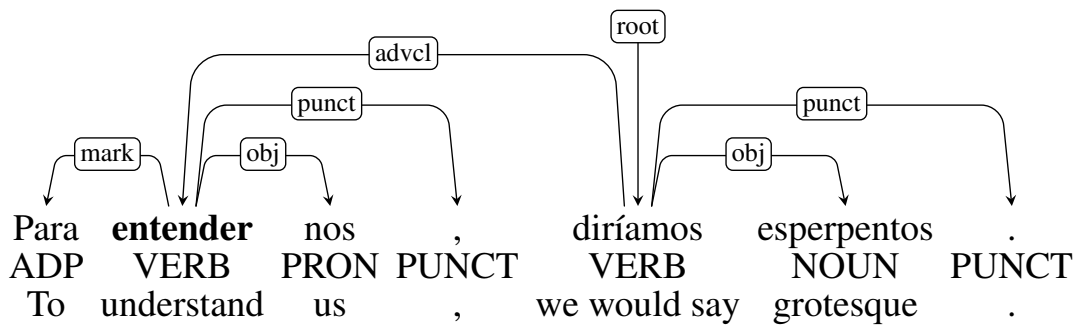

Figure 15: An example of `ArgM:fin` – advcl.

FGD serves as a basis for the Prague Dependency Treebank (Hajič et al., 2006; Bejček et al., 2013) and its successors such as Prague Czech-English Dependency Treebank (Hajič et al., 2012), the PDT of Spoken Czech (Mikulová et al., 2017) and the Czech-English Parallel Corpus (Bojar et al., 2011; Mareček, 2011).

In other existing resources that adopted the PDT annotation scheme, the tectogrammatical layer does not seem available, with the exception of the Index Thomisticus treebank of Latin (Passarotti, 2014), whose deep-syntactic annotation is close to PDT.

### 4.2 Available data

The Prague Dependency Treebank (PDT) is a treebank consisting of a subset of the Czech National Corpus. Its domain is mainly newspaper texts and business and popular scientific articles from the 1990s. The deep-syntactic annotation is available in the PDT from version 2.0 onward (Hajič et al., 2006).

The Prague Czech-English Dependency Treebank (PCEDT) is a manually annotated parallel, aligned treebank that contains the Wall Street Journal text collection of the Penn Treebank. The English part of PCEDT 2.0 contains the entire Penn Treebank – Wall Street Journal Section. The Czech part consists of Czech translations of all of the Penn Treebank-WSJ texts. The corpus is 1:1 sentence-aligned. An additional automatic alignment on the node level (different for each annotation layer) is available as well (Hajič et al., 2012).

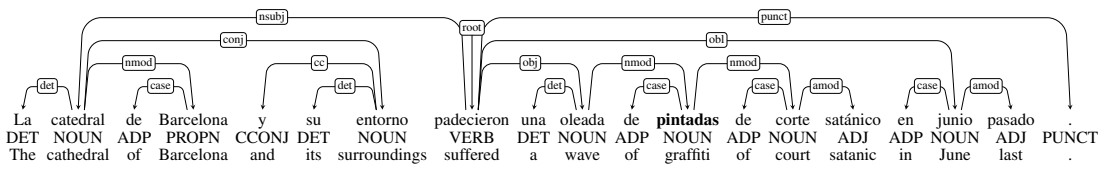

Figure 16: An example of `Arg0:src` – nmod.

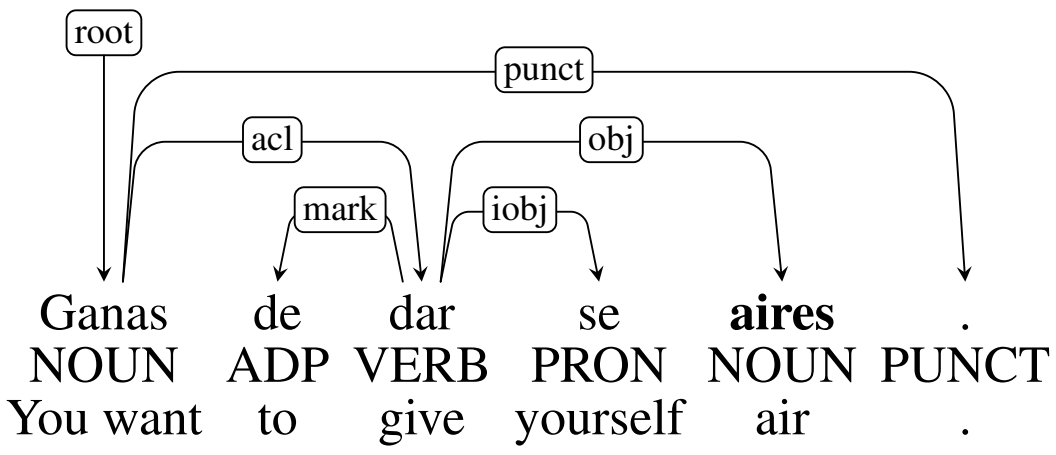

Figure 17: An example of `ArgL:` – obj.

As mentioned in Section 2, this proposal is oriented towards UD. For that reason, the Czech-PDT UD treebank was used to study the data. The Czech-PDT UD treebank is based on the Prague Dependency Treebank 3.0 (PDT). Although the original semantic role labels (functors) are not available in the UD data, it was possible to automatically transfer them from the original PDT data using a Python script.[7] The same method was applied to the UD version of the Prague Czech-English Dependency Treebank (PCEDT)[8] (Figure 18).

Table 3 shows the number of sentences and tokens for each corpus.

Table 4 shows 67 semantic role labels (functors)[9] that were found in the P(CE)DT data and their frequencies

### 4.3 Semantic Role Labels

Semantic roles (functors) are divided into arguments and adjuncts according to both semantic and formal criteria specified within the valency theory (Panevová, 1974).

FGD specifies five argument roles that correspond mostly to the surface-syntactic slots of a subject and of direct and indirect objects of the verb.

**ACT** argument: actor

**ADDR** argument: addressee

**EFF** argument: effect

**ORIG** argument: origo

**PAT** argument: patient

Other types of verbal modifications are considered adjuncts. These functors corresponds to temporal, locational, manner and other kinds of adverbials. They can be classified by their intended purpose.

Functors for the effective roots of independent clauses: express the independence of the given lexical unit and determine the clause type.

**DENOM** independent nominal

**PAR** parenthetic clause

**PARTL** independent interjection

**PRED** independent verbal clause

**VOCAT** independent vocative

Temporal functors express various temporal points or intervals that the content of a governing modification relates to. Individual temporal functors differ according to which of the possible questions about time they answer.

**TFHL** adjunct: for how long

**TFRWH** adjunct: from when

**THL** adjunct: (after) how long

**THO** adjunct: how often

**TOWH** adjunct: to when

**TPAR** adjunct: in parallel with what

**TSIN** adjunct: since when

**TTILL** adjunct: until when

**TWHEN** adjunct: when

Locative and directional functors express location or direction related to the content of the governing word. The individual functors differ according to the kind of question they answer.

[7] The original PDT functors will be available in the Czech-PDT UD treebank starting from v2.12.

[8] Access to the UD version of the PCEDT data is restricted due to licensing restrictions. The original Prague Czech-English Dependency Treebank is available in the LDC catalog.

[9] Three extra labels were found in PCEDT: NE (named entity): 37,369, DESCR (adnominal description): 4,077, and SM: 508. However, they were not mentioned in the original PDT guidelines (Mikulová et al., 2006).

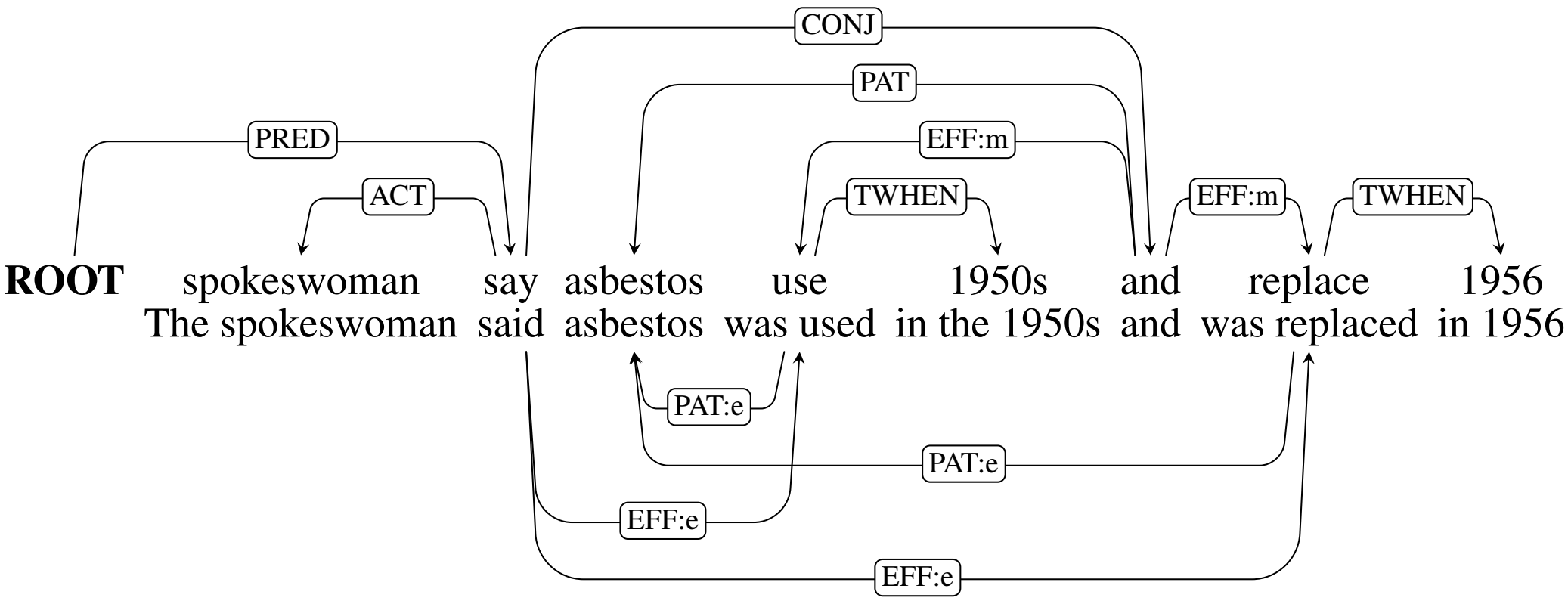

Figure 18: Propagation of effective dependencies to a shared argument of coordinate verbs. The sentence is *The spokeswoman said asbestos was used in the 1950s and replaced in 1956.* Note that the auxiliary *was* occurs only once but is included in anchoring of two nodes.

| Set | UD_Czech-PDT, sent./tok. | UD_English-PCEDT, sent./tok. |
|---|---|---|
| Train | 38,725/672,065 | 39,832/950,028 |
| Development | 5,228/90,813 | 6,960/167,054 |
| Testing | 5,475/95,758 | 2,416/56,684 |

Table 3: **UD_Czech-PDT, sent./tok.:** the number of sentences/tokens in the UD Czech-PDT treebank (only the part that has tectogrammatical annotation is counted); **UD_English-PCEDT, sent./tok.:** the number of sentences/tokens in the UD English-PCEDT treebank.

**DIR1** adjunct: where from

**DIR2** adjunct: which way

**DIR3** adjunct: where to

**LOC** adjunct: where

Functors for causal relations express various implicational (causal) relations between events or states. The choice of the functor reflects the type of the relation between these two events or phenomena (cause, condition, purpose, or concession etc.)

**AIM** adjunct: purpose

**CAUS** adjunct: cause

**CNCS** adjunct: concession

**COND** adjunct: condition

**INTT** adjunct: intention

Functors for expressing manner constitute a broad category of adjuncts that express all kinds of inner characteristics of events – the manner in which the event (state) proceeds/comes about. For instance, by comparison, by specifying the result or instrument used for accomplishing the event, by expressing quantity etc.

**ACMP** adjunct: accompaniment

**CPR** adjunct: comparison

**CRIT** adjunct: criterion

**DIFF** adjunct: difference

**EXT** adjunct: extent

**MANN** adjunct: manner

**MEANS** adjunct: means

**REG** adjunct: with regard to

**RESL** adjunct: result

**RESTR** adjunct: exception, restriction

The **COMPL** functor is assigned to predicative complements (i.e. optional adjuncts with a dual semantic relation).

The **CM** functor is a functor assigned to conjunction modifiers, mostly various particles and adverbs.

Functors for specific modifications are assigned to certain specific modifications that are not traditionally included in the syntactic descriptions. These functors could belong to the group of functors for manner and its specific variants.

**BEN** adjunct: benefactor

**CONTRD** adjunct: confrontation

**HER** adjunct: inheritance

**SUBS** adjunct: substitution

Specific adnominal functors are designed exclusively for modifying (semantic) nouns. The verbal functors only are not sufficient for representing all the functions of adnominal modifications.

**APP** adnominal adjunct: appurtenance

**AUTH** adnominal adjunct: author

**ID** adnominal specification of identity

**MAT** adnominal argument: content

**RSTR** adnominal adjunct: modification

Functors for rhematizers, sentence, linking and modal adverbial expressions are designed for representing free modifications and their function in the sentence – to rhematize, to link the sentence to its preceding context or to express various modal meanings and attitude.

**ATT** speaker's attitude

**INTF** expletive subject

**MOD** some modal expressions

**PREC** preceding context

**RHEM** rhematizer

Functors for multi-word lexical units and foreign-language expressions are used for representing certain multi-word lexical units or foreign-language parts that are not strictly analyzed.

**CPHR** nominal part of complex predicate

**DPHR** dependent part of idiom

**FPHR** part of foreign expression

Functors expressing the relations between the members of paratactic structures do not comply with the general definition of a functor as a semantic value of the syntactic relation of dependency. They do not express a kind of dependency; they rather capture the relation between members of paratactic structures (either clauses or modifications).

**ADVS** parataxis: adversative

**APPS** parataxis: apposition

**CONFR** parataxis: confrontation

**CONJ** parataxis: conjunction

**CONTRA** parataxis: conflict

**CSQ** parataxis: consequence

**DISJ** parataxis: disjunction

**GRAD** parataxis: gradation

**OPER** parataxis: math operation

**REAS** parataxis: cause

A limited group of functors can be further subclassified using subfunctors. Subfunctors describe semantic variation within a particular functor, typically these differences are expressed by various prepositional phrases, by using different cases or conjuctions.

Subfunctors are not included in this research – this level of granularity goes against the goal of creating a unified set of semantic role labels.

## 5 Universal Semantic Role Labels

The proposed set of unified semantic labels consists of 14 labels. A unified semantic role label is structured as follows *MAIN:subcategory*.

The main label expresses the main semantic category. Some of the labels do not hold any semantic value, they rather capture the relation between members of paratactic structures or other units. Although the approach does not lean towards a valency dictionary, the distinction between arguments and adjuncts is preserved.

The sub-categorical part of the label is not obligatory. It is designed for preserving original semantic information that may be available in the data.

**Arguments**

As for the argument labels, the original FGD approach is preserved, with the intention to assign an identical deep representation to an active and passive diathesis of the same predicate.

**ACT** - corresponds to the ACT functor, the Agent role label **agt** (Section 3.3), the Cause role label **cau** (Section 3.3) when it is associated with the subject (Section 3.3), and the Experiencer role label **exp** (Section 3.3). This label can have the following subcategories:

**ACT:agt**
**ACT:cau**
**ACT:exp**

**ADDR** - refers to the Addressee of an event

**EFF** - refers to the result of the event

**ORIG** - refers to the origin/source of an event; corresponds to the ORIG functor, and can be extended as **ORIG:src**. The extension corresponds to the Source role label **src** (Section 3.3)

**PAT** - corresponds to the PAT functor, the Patient role label **pat** (Section 3.3), the Theme role label **tem** (Section 3.3) when it is expressed as `arg1`. This label can have the following subcategories:

**PAT:theme**

**PAT:atr** that corresponds to the Attribute role label **atr** (Section 3.3)

**Manner: MANR**

The **MANR** label refers to adjuncts of manner that describe how the action, experience, or process of an event is carried out. This label combines the functors for expressing manner, the functors for specific modifications, the Manner role label **mnr** (Section 3.3), and the Beneficiary role label **ben** (Section 3.3) enabling the following combinations if the additional information is available:

**MANR:acmp**

**MANR:cpr**

**MANR:crit**

**MANR:diff**

**MANR:ext**

**MANR:mann** - corresponds to the MANN functor and the mnr role label

**MANR:means**

**MANR:reg**

**MANR:resl**

**MANR:restr**

**MANR:ben** - corresponds to the BENN functor and the ben role label [10]

**MANR:contrd**

**MANR:her**

**MANR:subs**

**Locative: LOC**

The **LOC** label is bound to location or direction. This label combines the Locative and directional functors, the Location role label **loc** (Section 3.3), the Destination role label **des** (Section 3.3), and the Origin role label **ori** (Section 3.3) enabling the following combinations if the additional information is available:

**LOC:dir1**

**LOC:dir2**

**LOC:dir3** - corresponds to the DIR3 functor and the des role label

**LOC:where** - corresponds to the LOC functor

textbfLOC:ori - corresponds to the ori role label

**Causal: CAUSE**

The CAUSE label refers to adjuncts that express various causal relations. It includes various functors from the causal group, the Cause role label **cau** (Section 3.3) when it takes an adjunct position, and the Purpose label **fin** (Section 3.3).

**CAUSE:aim** - corresponds to the AIM functor and the fin role label

**CAUSE:caus** - corresponds to the CAUS functor and the cau role label

[10] There might be language-specific cases when it would be difficult to distinguish between the Benefactor and Addressee roles.

**CAUSE:cncs**

**CAUSE:cond**

**CAUSE:intt**

**Temporal: TIME**

The **TIME** label refers to temporal adjuncts that express various temporal points or intervals. This label combines the Temporal group of functors and the Time role label **tmp** (Section 3.3) enabling the following combinations if the additional information is available:

**TIME:fhl**

**TIME:frwh**

**TIME:hl**

**TIME:ho**

**TIME:owh**

**TIME:par**

**TIME:sin**

**TIME:till**

**TIME:when**

**Paratactic: BINDER**

The **BINDER** label refers to paratactic structures and captures the relation between different parts of the utterance.

**BINDER:advs**

**BINDER:apps**

**BINDER:confr**

**BINDER:conj**

**BINDER:contra**

**BINDER:csq**

**BINDER:disj**

**BINDER:grad**

**BINDER:oper**

**BINDER:reas**

**Independent Clauses: IND**

The **IND** label is designed for the functors that express the independence of the given lexical unit and determine the clause type.

**IND:denom**

**IND:par**

**IND:partl**

**IND:pred**

**IND:vocat**

**Predicative complement: PCOMPL**

The **PCOMPL** label is designed for adjuncts that are expressed as predicative complements.

This label combines the COMPL functors, the Final State role **efi** (Section 3.3), and the Initial State role **ein** (Section 3.3) enabling the following combinations if the additional information is available:

**PCOMPL:compl**

**PCOMPL:efi**

**PCOMPL:ein**

**Adnominal: ADNOM**

Since there is no counterpart for this case in the MTT AnCora label sat, the **ADNOM** label is designed for functors that are assigned to modifications exclusively modifying (semantic) nouns.

**ADNOM:auth**

**ADNOM:id**

**ADNOM:mat**

**ADNOM:restr**

**Miscellaneous: MISCLL**

The **MISCLL** label is designed for miscellaneous relations. It combines two groups of functors: the functors for rhematizers, sentence, linking and modal adverbial expressions and the functors for multi-word lexical units and foreign-language expressions.

**MISCLL:att**

**MISCLL:cm**

**MISCLL:cphr**

**MISCLL:dphr**

**MISCLL:fphr**

**MISCLL:intf**

**MISCLL:mod**

**MISCLL:prec**

**MISCLL:rhem**

## 6 Conclusion

We have surveyed the label inventories of deep-syntactic relations from two theories and annotation frameworks: Meaning-Text Theory and Functional Generative Description. Based on the observations, we proposed a unified relation inventory, which contains unified labels for relations that are similar or equivalent in the two frameworks, and additional labels for relations that are unique, so that annotations can be mapped with minimal information loss. The unified inventory is hierarchical so that less-specific relation types can be mapped and the missing finer distinctions do not have to be guessed.

In the future, we intend to add mapping from other annotation schemes, such as Sequoia or PropBank. It is possible that the universal relation set will have to be slightly modified as a result; however, the two current source frameworks (and in particular FGD) have quite detailed inventories of relations, therefore we believe that the proposed universal set already covers a substantial part of what can be found in deep-syntactic datasets in general.

# A Appendix: Semantic role label frequency in the PDT and PCEDT data.

Table 4: Semantic role label frequency in the PDT and PCEDT data.

| Label | Czech | English |
|---|---|---|
| RSTR | 142,509 | 161,142 |
| ACT | 94,644 | 127,884 |
| PAT | 89,248 | 127,173 |
| PRED | 50,472 | 53,352 |
| APP | 29,807 | 32,446 |
| CONJ | 24,640 | 22,536 |
| LOC | 21,980 | 23,804 |
| TWHEN | 19,336 | 25,723 |
| RHEM | 12,320 | 10,104 |
| MANN | 9,784 | 7,587 |
| EXT | 8,302 | 11,564 |
| EFF | 7,704 | 21,044 |
| ADDR | 6,761 | 8,323 |
| PREC | 6,126 | 5,697 |
| DIR3 | 6,075 | 4,455 |
| DENOM | 5,877 | 1,296 |
| ID | 5,808 | 1,971 |
| BEN | 5,540 | 3,801 |
| MAT | 5,210 | 5,047 |
| DIR1 | 4,927 | 3,766 |
| APPS | 4,461 | 10,284 |
| FPHR | 4,192 | 2,484 |

Continued on next page

Table 4: Semantic role label frequency in the PDT and PCEDT data. (Continued)

| | | |
|---|---|---|
| PAR | 4,191 | 3,124 |
| ACMP | 3,780 | 3,404 |
| REG | 3,165 | 12,211 |
| CM | 2,958 | 1,248 |
| MEANS | 2,870 | 2,215 |
| CPHR | 2,707 | 3,138 |
| CAUS | 2,612 | 3,293 |
| COND | 2,525 | 2,101 |
| CRIT | 2, 485 | 1,464 |
| AIM | 2,415 | 4,991 |
| THL | 2,105 | 2,582 |
| ADVS | 2,048 | 2,105 |
| ATT | 2,026 | 1,587 |
| MOD | 1,770 | 738 |
| COMPL | 1,558 | 3,522 |
| OPER | 1,538 | 984 |
| THO | 1,401 | 1,935 |
| DPHR | 1,263 | 896 |
| DISJ | 1,254 | 1,947 |
| TTILL | 1,219 | 1,212 |
| ORIG | 1,095 | 3,617 |
| DIFF | 931 | 5,007 |
| TSIN | 924 | 962 |
| CPR | 885 | 2,532 |
| AUTH | 774 | 449 |
| CNCS | 753 | 1,056 |
| RESTR | 738 | 744 |
| GRAD | 709 | 226 |
| TPAR | 608 | 1,514 |
| CSQ | 470 | 205 |
| DIR2 | 447 | 296 |
| RESL | 399 | 641 |
| TFHL | 375 | 1,300 |
| SUBS | 303 | 321 |
| REAS | 243 | 68 |
| INTT | 189 | 127 |

Continued on next page

Table 4: Semantic role label frequency in the PDT and PCEDT data. (Continued)

| | | |
|---|---|---|
| TOWH | 185 | 153 |
| CONTRD | 168 | 547 |
| CONTRA | 168 | 56 |
| TFRWH | 166 | 431 |
| INTF | 126 | 8 |
| PARTL | 102 | 106 |
| VOCAT | 60 | 31 |
| HER | 33 | 13 |
| CONFR | 23 | 75 |

## B Appendix: Thematic role label frequency in the AnCora data.

| **Label** | **Spanish** | **Catalan** | **Associated arg position** |
|---|---|---|---|
| Patient: pat | 29922 | 24039 | arg1 |
| Agent: agt | 24951 | 18432 | arg0 |
| Theme: tem | 23628 | 18318 | arg1, arg2, arg3 |
| Attribute: atr | 11864 | 10616 | arg2, argM, arg3 |
| Time: tmp | 10674 | 9014 | argM |
| Location: loc | 9406 | 8057 | argM, arg2, arg1, arg3 |
| Adverbial: adv | 9852 | 6581 | argM |
| Cause: cau | 3627 | 2726 | argM, arg0 |
| Beneficiary: ben | 3027 | 2112 | arg2, arg3 |
| Purpose: fin | 2203 | 1788 | argM, arg2 |
| Manner: mnr | 1945 | 1669 | argM |
| Final State: efi | 684 | 596 | arg2, arg4 |
| Destination: des | 862 | 523 | arg4 |
| Empty label: ” | 725 | 511 | argL |
| Extension: ext | 725 | 472 | arg2, argM, arg1 |
| Cotheme: cot | 148 | 251 | arg2, arg1 |
| Origin: ori | 386 | 198 | arg3 |
| Experiencer: exp | 174 | 125 | arg2, arg3, arg0 |
| Initial State: ein | 112 | 109 | arg3, arg2 |
| Instrument: ins | 88 | 36 | arg2 |
| Source: src | 30 | 4 | arg0 |

Table 5